\documentclass{article}
\pdfoutput=1
\usepackage{geometry}
\usepackage{amsmath,amssymb}
\usepackage{natbib}
\usepackage{graphicx}
\usepackage{url}
\usepackage{algorithm2e}
\usepackage{nameref}
\usepackage{authblk}

\title{A discriminative approach for finding and characterizing positivity violations using decision trees}

\author{Ehud Karavani}
\author{Peter Bak}
\author{Yishai Shimoni\footnote{Corresponding author: yishais@il.ibm.com}}
\affil{IBM-Research Haifa Labs}

\date{}

\begin{document}

\maketitle

\begin{abstract}
The assumption of positivity in causal inference (also known as common support and covariate overlap) is necessary to obtain valid causal estimates. 
Therefore, confirming it holds in a given dataset is an important first step of any causal analysis.
Most common methods to date are insufficient for discovering nonpositivity, as
they do not scale for modern high-dimensional covariate spaces, or they cannot pinpoint the subpopulation violating positivity. 

To overcome these issues, we suggest to harness decision trees for detecting violations.
By dividing the covariate space into mutually exclusive regions, each with maximized homogeneity of treatment groups, decision trees can be used to automatically detect subspaces violating positivity.
By augmenting the method with an additional random forest model, we can quantify the robustness of the violation within each subspace. 
This solution is scalable and provides an interpretable characterization of the subspaces in which violations occur.


We provide a visualization of the stratification rules that define each subpopulation, combined with the severity of positivity violation within it. 
We also provide an interactive version of the visualization that allows a deeper dive into the properties of each subspace.


\end{abstract}

\section{Introduction}

Causal inference is the field dedicated to estimating causal effects of some interventions from real world retrospective data. 
At its most basic concept, estimating an effect of an intervention is done by comparing two groups, one that received the intervention, and another that received a different intervention (e.g., placebo or no intervention).
However, there are three assumptions needed for this derived quantity to be interpreted as a \emph{causal} effect, 
namely, no unmeasured confounders, consistency, and positivity \citep{hernan2010causal}.

Positivity is the assumption that every sample has some positive probability to be assigned to every treatment. 
If we denote covariate space as $X$ and treatment assignment as $A$, then
positivity can be formalized as $\Pr[A=a|X]  \forall a \in A$ \citep{westreich2010invited}. 

Violation of this assumption, also known as non-positivity, or lack of covariate overlap, 
suggests that there is some combination of features (an instantiation of $X$) for which the probability to be treated is either 1 or 0, and hence, deterministic. 
This implies that that particular subspace of covariates is populated solely by samples from one group. 
This, in turn, makes it impossible (or much less probable) to infer the counterfactual outcome for these samples,
as there is no data from the other group to infer from.

In some cases, such as when caused by randomness, non-positivity might be solved by some interpolation. 
In others, the cause can be structural (for example, medical guidelines prevent people over 60 to be assigned a treatment). 
In such cases, cohort refinement is necessary, exclusion criteria must be updated, 
and the causal question needs to be adjusted \citep{westreich2010invited} 
(for example, if all people over 60 are excluded then the estimand is no longer 
“effect in the population”, but rather “effect in people under 60”). 
Ignoring the violation invalidates the causal question as it makes the two groups incomparable. 

Therefore, it is of interest for researchers to be able to easily identify if violations exist in the data, 
and moreover, to characterize the covariate subspaces in which violations occur. 
Nevertheless, as the complexity of covariate spaces increases, 
by being of both high-dimensional and of various types (i.e., mixing continuous, binary and categorical variables), 
this task becomes highly challenging. 

\subsection{Related work}
There are few methodologies currently in practice to find violations of positivity. 
These can be broadly cast into two groups, 
one that is based on statistical tests comparing the two groups on each covariate marginally, 
and another based on comparing propensity scores across these two groups.

\subsubsection{Marginally comparing covariates}
The task of finding lack of covariate overlap can be translated to the task of checking whether the distribution of covariates is similar between the two treatment groups. 
Checking for distances between distributions is a thoroughly investigated problem in statistics. 
However, most statistical tests do not scale naturally for multivariate distributions \citep{biswas2014nonparametric}. 
Additionally, multivariate density functions cannot be assessed easily to help observe how distributions of the two groups differ.
 Therefore, an intuitive solution is to reduce the multivariate task to many single-variate tasks.
In other words, examining the distribution of each covariate separately (i.e., marginally) \citep{linden2015graphical}. 
Note, however, that overlap in each covariate marginally does not guarantee the multivariate joint distribution also overlap, as exemplified in Figure \ref{fig:fig1}A, 
showing a synthetic 2-dimensional covariate space in which positivity is clearly violated, but this cannot be observed in any single dimension.

To overcome this generalization of marginals to joint overlap, \cite{messer2010effects} suggested tabular analysis to check for violations on the intersection of several covariates. 
While their method characterizes the violating subspaces in a simple and intuitive way, 
it has two drawbacks: 
(1) it does not scale properly for more than a handful of covariates, due to its 2-dimensional tabular nature; and
(2) the handling of continuous variables is suboptimal, as it is done by using predefined bins (e.g., quartiles).
Such binning can be intuitive, but also quite sensitive, since the number of bins (as well as the spacing of their edges) can significantly impact the false-positive and false-negative detection rates. 
In addition, proper binning is particularly hard for heavy-tail distributions that are abundant in real-world data.

\subsubsection{Propensity based methods}
Another approach to mitigate this "curse of dimensionality" is with the use of propensity scores (the probability $\Pr[A=1|X]$). 
From a theoretical perspective \citep{rosenbaum1983central}, it is possible to compare the distribution of propensity scores (rather than covariates)
 between groups and alert for violations once the mass of the two distributions does not overlap, especially their tails (see Figure \ref{fig:suppfig2}).

This method attracts researchers for being grounded in theory. However, it also has two main drawbacks. 
Firstly, propensity scores can be viewed as a dimensionality reduction from the covariate space onto a single number, 
and almost always bear some information loss \citep{geiger2012relative}. 
Specifically, two different subspaces can be mapped into the same scalar, 
suggesting an observed overlap can picture an overly optimistic view of the true violations in the original covariate space. 
Secondly, once a violation is detected, it is usually hard to characterize the covariate subspace causing it, 
as it depends on the interpretability of the model used to obtain those propensity scores.
Therefore, while this approach can provide an overview of the problem, it usually fails to convey information beyond the binary  “a problem does/does-not exists”.

\paragraph{In this work,} we introduce a method to identify whether violations of positivity exist in a given dataset. 
While current methods seem to trade-off between the two, our method can both scale for complex high-dimensional spaces, while also being highly interpretable.

\section{Results}

As its name suggests, lack of common support can be detected by measuring large statistical distance between the covariate distribution of treated and untreated groups. 
This comparison of distributions is known in statistics as \emph{two-sample test} \citep{lehmann2006testing}.
To accommodate these tests for complex high-dimensional spaces, recent works have suggested utilizing automatic machine learning discriminators to try and differentiate between the two samples 
\citep{lopez2016revisiting,ozery2018adversarial,goodfellow2014generative}.

Inspired by this approach, we apply a machine learning model to detect positivity violations.
Essentially, if a violation exists (i.e., common support is lacking), it suggests a volume of the covariate space is occupied solely by one group. 
This subspace can then be identified by the model, which will use the samples populating it to achieve better discrimination score.
Ideally, on well-balanced datasets, like the ones needed for causal inference, the model will not be able to discriminate the two groups and its prediction will be no better than a random coin flip.

However, we do not use any discriminitaive model, but decision trees specifically. 
Decision trees are hierarchical classification models, usually constructed by greedy algorithms. 
At each step, the algorithm selects one covariate and one cutoff-value, so when splitting on them, the resulting subsets have maximal homogeneity \citep{quinlan1986induction}.
This process results in the decision tree dividing the entire covariate space into mutually exclusive regions such that each region has its purity (i.e., group homogeneity) maximized \citep{rokach2005decision}. 



We harness this property for the detection of positivity violations.
We train a decision tree classifier using the treatment assignment as the target variable.
While being constructed, the tree tries to maximize group discrimination by finding regions in the covariate space populated exclusively by only one treatment group.
These covariate subspaces are, by definition, violating the positivity assumption.



To characterize the subspaces violating positivity, we utilize another useful property of decision trees which is their interpretability.
Since the covariate space has been divided into disjoint union of regions, it enables us to refer to covariate subspaces and decision tree's leaves interchangeably \citep{rokach2005decision}.
Furthermore, since the division is done by simple stratification rules, 
it allows translating a path from root to leaf into a covariate-level query and provide a simple and intuitive characterization of the samples found violating positivity (see \nameref{methods}).

To illustrate, once a tree is constructed, we go over its leaves and examine their purity. 
If a leaf contains samples from both the treated and untreated groups, we can assume that these samples are not violating positivity. 
But if a leaf contains samples from only one group and not the other (i.e., a homogenic leaf),
we can assume that the subspace defined by the leaf is violating positivity, and the samples mapped to that leaf should be handled.

It is worth noting, however, that decision trees are notorious for overfitting. 
With unlimited depth, they can divide the space so much that each datapoint will be classified correctly \citep{rokach2008data}. 
This implies they can be overly-sensitive to violations, not all of which are meaningful, as many violating subspaces will be comprised from a handful of samples each.
One can be easily convinced that this overly-fine division of the space is similar to the depiction of decision tree's overfit. 
Therefore, controlling for the tree's overfit also controls for the sensitivity of violation detection.
Luckily, overfitting can be mitigated automatically by constructing a heavily, yet carefully, regularized tree; one that will allow the model to generalize well to new out-of-bag samples \citep{kohavi1995study} (see \nameref{methods}). 
Nonetheless, reducing the rates of false-positives might come at the price of increasing the rate of true violations the tree falsely failed to detect (i.e., higher false-negatives).
Alternatively, hyper-sensitivity might be seen as an advantage for researchers. 
As it will allow them to go over the putative violations flagged by the algorithm and apply domain-expert knowledge to understand whether they are meaningful or not, while knowing they are less likely to miss true violations due to the model's low false-negative rate.

Likewise, we can also hint at whether a violation is due to randomness or not, by using a random forest. 
Random forests are ensembles of decision trees, originally invented to mitigate the tendency of a single decision tree to overfit, a problem we just linked to the meaningfulness of putative violations. 
They do so by constructing many trees, each using a bootstrap sample of the dataset (a random sample with replacement) and a smaller random set of covariates, and then aggregating each tree’s prediction into a final one \citep{breiman1984classification}. 

We use a random forest to test the consistency of a datapoint being tagged as a violation, by applying all trees to the original dataset and counting the fraction of trees in which a sample was considered in violation. 
If that number is small, we can suspect the violation was due to a random "fluke" in the dataset (as the forest itself builds on a random sample of the dataset) captured by the tree’s overfit. 
This logic is unidirectional, since if the fraction is large, we cannot conclude the violation has a structural cause, since it can still arise from a sampling error in the original dataset. 
It is up to the researcher to investigate the subspace in order to determine if a consistent violation is due to structural causes or randomness, and how to resolve it.

Lastly, this approach allows us to control for what is defined as a positivity violation. 
Strictly speaking, a violation is a covariate subspace $x$ which contains samples from one group only, which translates to $\Pr[A=a|X=x]$ equaling either 1, or 0. 
But consider the case that, for example, of having 10,000 untreated samples over the age of 60, and only 1 treated sample. 
We define those cases as \emph{soft violations}. 
Whether these should be considered a violation or not are left to the discretion of the researcher, as it translates to a bias-variance tradeoff; 
since excluding those samples will bias the original estimand but including them will cause the estimator to be highly variable. 
To illustrate, from inverse probability weighting perspective, that single treated individual now counts as 10,000 ones.
Leaf homogeneity is directly linked to how balanced the corresponding subspace is,
and by tuning the threshold of leaf homogeneity, we can define what counts as violation (see \nameref{methods}).

\subsection{Use case}
\begin{figure}[h]
  \includegraphics[width=\linewidth]{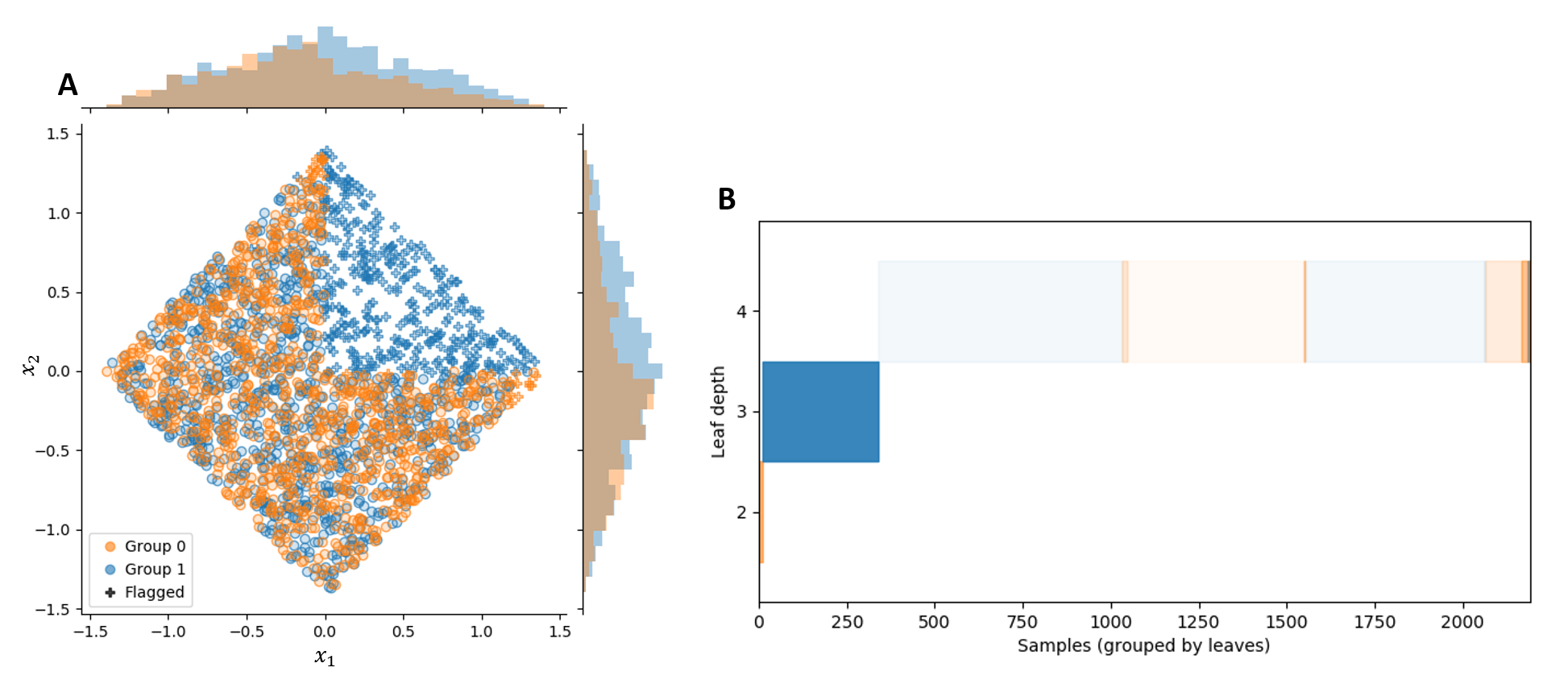}
  \caption{A synthetic example of applying positivity-detection tree. (\textbf{A}) Scatter plot of the two-dimensional data containing a positivity violation, where samples in the first quadrant are exclusively assigned to group 1 (blue) and not group 0 (orange). Samples that where flagged as being a part of a violating subspace are depicted in 'plus' marker. Marginal distributions are also presented (top and right histograms) to show that examining each covariate separately would not detect a violation exist. (\textbf{B}) Visualization of the corresponding tree. Main observation is that the tree was able to capture the violation in the first quadrant (prominent rectangle in corresponding blue). We note another opaque orange box to its left that the tree was able to detect, but being slim means the number of samples is small and therefore can be negligible. The rest of the rectangles are transparent, suggesting that the individuals are not participating in violating subspaces in a consistent way.}
  \label{fig:fig1}
\end{figure}

\begin{figure}[h]
  \includegraphics[width=\linewidth]{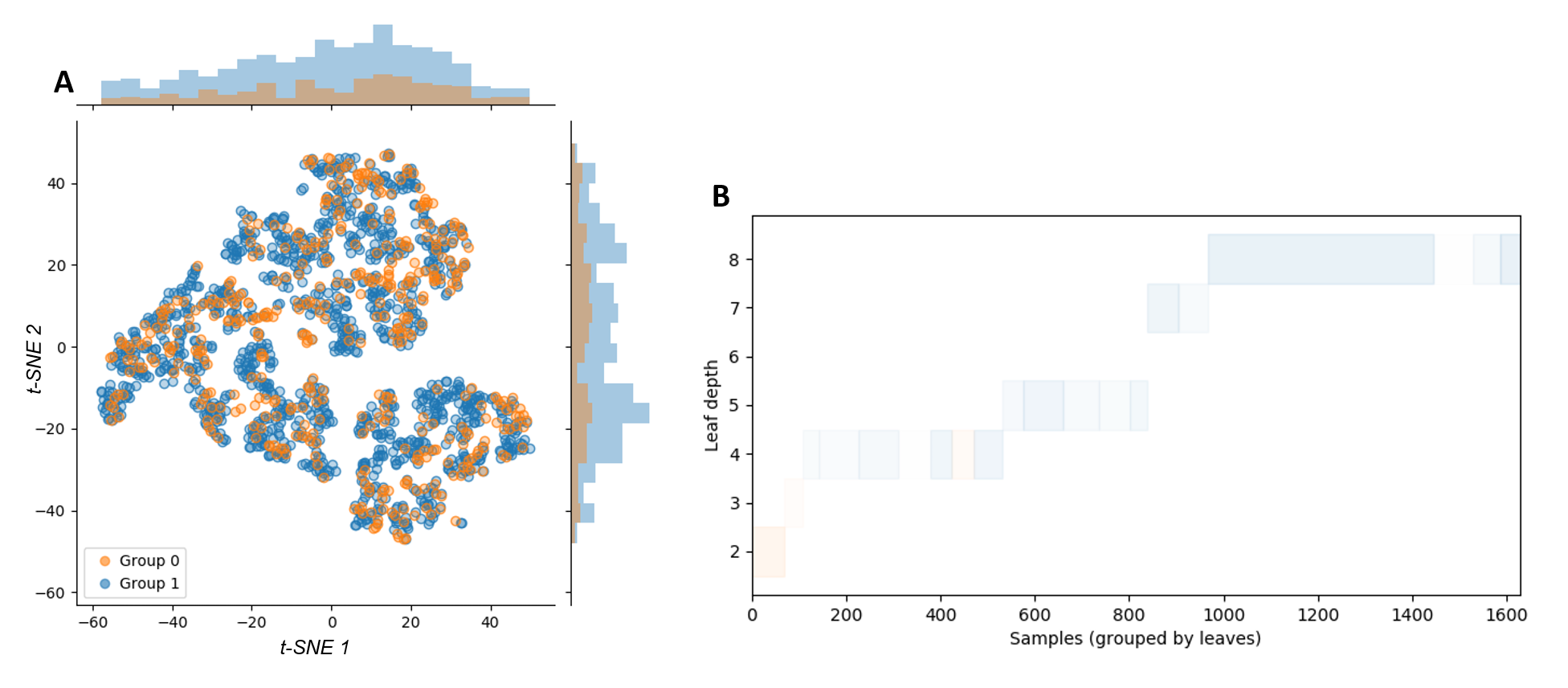}
  \caption{Similar to Figure 1 but using real data from the NHEFS studying the effect of smoking cessation on weight gain. Since the data is of high-dimension, (\textbf{A}) contains a scatter plot of the t-SNE projection onto two-dimensions of the data. Overlap of points from group 0 (orange) and group 1 (blue) hints there are no major violations of nonpositivity (as close points in the projected space are also close with high probability in the original space). Applying our method on the original domain space and visualizing it in (\textbf{B}) confirms this, as the rectangles are so transparent (meaning, inconsistent in their units violating positivity), that the plot seems to be blank.}
  \label{fig:fig2}
\end{figure}

A main component of our method is its visualization. 
Briefly, we visualize the leaves of our decision-tree as fixed-height rectangles with their y-axis location corresponding to tree-depth and their width corresponding to the number of samples mapped to them (see \nameref{methods} and Figures \ref{fig:fig1}B, \ref{fig:fig2}B). 
Each treatment group is assigned a categorical color and rectangles are colored by the majority group of the each leaf.
Additional color opacity is controlled by the consistency measurement derived from the forest.

Size and coloring are incorporated to give an easy overview on the severity of the violations found in the dataset.
Large violating subspaces (as in many samples belonging to them) will occupy larger real-estate, and
consistent violations (repeatedly rediscovered in the random forest) will be opaque and intense.
The combination of the two ensures that meaningful violations are visualized conspicuously.
Moreover, an interactive version (Figure \ref{fig:suppfig1}), allows the researcher to hover over the tree to get more details on the violation (e.g., exact number of samples from each group, the query to extract that subspace, etc.).

We present two use cases for the method. The first one, shown in Figure \ref{fig:fig1}, is a synthetic example of a rotated square. It shows two groups (orange and blue), overlapping except for the first quadrant having samples exclusively from group 1, suggesting $\Pr[A=1|X_1>0, X_2>0]=1$. The marginal distributions of the two groups are similar and would avoid detection by the marginal methods described above. Comparing propensity scores by group (Figure \ref{fig:suppfig2}), would detect a problem, but reverse engineering these propensities to characterizable subspaces is model dependent. 

Figure \ref{fig:fig1}B shows the decision-tree corresponding to the data from panel A. 
The prominent blue rectangle corresponds to the violated subspace of the first quadrant. 
It is meaningful as it is both wide, opaque (see \nameref{methods}) and has low vertical location. 
In contrast, there’s an orange rectangle at the bottom left, but it is thin, and therefore contains few samples. 
Additional rectangles are transparent, suggesting they are not meaningful, as they were rarely found to violate positivity within the random forest (i.e., inconsistent).

Figure \ref{fig:fig2}, shows a real world example of an observational study about the effect of smoking cessation on weight gain \citep{nhefs}. 
Since this is a high-dimensional example, Figure \ref{fig:fig2}A presents a t-distributed stochastic neighbor embedding (t-SNE) projection \citep{maaten2008visualizing} of that data onto two dimensions. 
One property of t-SNE is that close points in the lower dimension are close in the original domain with high probability. 
We can see that the two groups overlap and that orange points are found next to blue ones, hinting no violation is present. 
Figure \ref{fig:fig2}B reinforces that, as we can see all leaves of the corresponding tree are transparent, indicating no discrimination could be made (in the original domain, of course), 
and so no positivity violation could be found. 

\section{Methods} \label{methods}

We used Python 3.6 together with scikit-learn \citep{pedregosa2011scikit} for the main analysis and tree construction. Matplotlib \citep{hunter2007matplotlib} was used for static tree visualization and Bokeh \citep{bokeh} for the interactive one. Code can be available by request, please contact the authors.

\subsection{Constructing a decision tree}
Decision trees can gain 100\% accuracy on any dataset, but that does not mean their prediction is good, as it will probably fail to generalize for out-of-box samples. Similarly, an overfitted tree is one that detects meaningless violations, since each sample can be assigned to a leaf of its own, creating a “violation” of 1 sample against 0 samples from the other group. 

Therefore, when building a decision-tree, it is of importance that it will be properly regularized. To perform this model selection, we used random hyperparameter search in a cross-validation fashion to find the hyperparameters that provided best performance (in terms of area under the receiver operating characteristic curve) on the validation folds. Other search paradigms can be plugged in the tool, as long as they adhere to scikit-learn’s interface.

\subsubsection{Constructing a forest}
When constructing a random forest, each tree in the forest should be specified similarly to the main reference tree. This is done to make all trees comparable, so insights drawn from the forest are relevant to the main decision tree.

\subsection{Significance testing}
There are several metrics we can apply to check for the significance of a violation.

\subsubsection{Consistency}
Consistency is a measure for samples. It is done by applying the forest on the dataset. We place an indicator for each tree and each sample indicating whether that sample was flagged for being a part of a violating subspace by that tree. 
The consistency of a sample is the fraction of times it was counted as a violation among all the trees in the forest (a number between zero and one). 
The consistency of a leaf is an aggregation of the consistency of the samples mapped to it. For each leaf in the main tree, the consistency of the samples mapped to that leaf is averaged.

\subsubsection{Probability}
Probability is a measure for leaves, which we model using hypergeometric distribution. 
Given two possible interventions $A={0,1}$, we can denote the number of samples from each group in the entire dataset as $N_0,N_1$ accordingly. 
For a given leaf (or a covariate subspace), we can have $n_0$ untreated and $n_1$ treated patients belong to it. 
These samples ($n_0,n_1$) originate from the root samples ($N_0,N_1$). 
Therefore, we can model the leaf as a sample from the population (root) the following way: 

\begin{itemize}
  \item $N=N_0+N_1$ – the population size,
  \item $K=N_1$ –  the number of “successes” in the population,
  \item $n=n_0+n_1$ – the number of draws,
  \item $k=n_1$  – the number of observed successes.
\end{itemize}

And the probability of a leaf is the probability mass function $\Pr[X=k]$ given $X \sim Hypergeometric(N,k,n)$. The smaller the probability the rarer it is to get that subspace by chance.

\subsubsection{Prediction performance}
Overall sense of the severity of the violations can be obtain by testing the prediction performance of the classification. This can be done by applying any of numerous classification metrics common in the field of machine learning \citep{hossin2015review}. Counterintuitively, keeping in mind that in contrast to usual machine learning, a desired score in our case is a one suggesting the classifier is no better than a random coin-flip (e.g., zero-one loss or area under the receiver operating curve of 0.5). Random-like prediction performance can hint that the covariates of the two intervention groups overlap and cannot be discriminated from each other.

\subsubsection{Soft violations}
Violations are calculated based on impurity measures – either entropy: $-\sum_i p_i \log p_i$, or gini: $1 - \sum_i p_i^2$ . 
The advantage of those impurity measures is that they are invariant to whether imbalance was caused due to $n_0 \gg n_1$ or $n_0 \gg n_1$. 
To obtain strict classical positivity, we would tune our tree to flag every leaf with impurity-measure of zero, which suggests either $n_0=0$ or $n_1=0$. 
However, as explained above, we might be interested in subspaces where unbalancing is simply large, i.e. the ratio $\frac{n_1}{n_0}$  is either very large or very small.
This can be obtained by setting some small positive threshold $t$, for which leaves with impurity below will be flagged as violating.

\paragraph{Relative violations}
A common case is that treatment prevalence is low in the overall population (i.e., $N_1<N_0$). 
In such cases, what regards as a soft violation should be adjusted according to the overall prevalence rate. 
Denoting the impurity measure $H$, we suggest comparing $H(root)$ to $H(leaf)$. 
Specifically, we calculate the condition $H(root)-H(leaf)>t'$, for a small positive $t'$, 
in order to determine whether to flag out a leaf. 
By enforcing positive difference (i.e. $H(leaf)>H(root)$) we assert that we don’t mind the leaf being more heterogenic than the overall population (the root). 
Given a desired threshold $t$ from above, it can be transformed into $t' = \max \{H(root)-t, 0\}$.

\subsection{Visualization}
We provide both static and interactive visualizations of the decision tree. The goal of the static visualization is to get a concise overview about how severe the violations in the dataset are. The interactive one can also further explore the leaves to obtain more details (as described above) about the subspaces they represent.

The visualization is built out of fixed-height rectangles. Each such rectangle represents a leaf, which width corresponding to the number of samples it contains and the y-axis location to its depth. 
Therefore, the entire of the x-axis represents all the samples in the datasets, grouped by their corresponding leaves; and shallower leaves in the tree are lower on the y-axis.

Each treatment group gets its own color. The color of a leaf is the color associated with the majority group mapped to it. Additional color opacity is determined by the average consistency values of the samples belonging to the leaf. The aggregation is not limited to average and could be also median or any other summary statistic over the consistency values.

Combining the shaping and coloring, we get that important subspaces are large (wide rectangle), consistent (opaque) and simple (few features involve in defining it, therefore the leaf is shallow in the tree, and therefore it is closer to the axes). On the other hand, unimportant subspaces will be depicted as transparent or small (negligible) rectangles. This way, the more eye-catching the plot is – the more severe violations are in the dataset, so ideally, one should expect a blank plot.

\subsection{Characterization - rule extraction}
Given a leaf, characterization of the subspace defined by it is straightforward. 

\begin{algorithm}[h]
    \SetAlgoLined
    \KwData{A tree $T$, a desired leaf $l$.}
    \KwResult{A covariate-level query $Q$ defining the subspace corresponding to $l$.}
    Initialize empty query $Q$.\;
   
    traverse from root node $T(root)$ to leaf $l$:

    \For{each node $k$ along the path}{
        	Let $X_i$ be the covariate inquired for split on node $k$\;

	Let $c$ be the cutoff for that split\;
	
	Let $s \in \{\le,>\}$ be the sign used to split\;

	Update $Q = Q \wedge (X_i s c)$\;
    }
\caption{Algorithm for extracting covariate-level characterization from a decision tree leaf, characterizing the samples belonging to the covariate subspace correspnding to that leaf.}
\end{algorithm}

Since a decision tree can reuse covariates in different depths of the tree, a subspace query can have redundant rules (e.g., $x_3>0 \wedge x_3>1$ that can be simply formatted as $x_3>1$). To simplify such repetitions, query pruning can be applied. When encountering a repeated rule (in terms of both covariate and cutoff sign), the rule closer to the leaf is kept, while the rule closer to the tree root is discarded. This promises rules kept are more specific to the subspace (as rules are more general the closer they are to the root).

\subsection{Data}
Real data is from a National Health Epidemiologic Followup Study (NHEFS) from the National Health and Nutrition Examination Survey (NHANES), studying the effect of smoking cessation on weight gain  \citep{nhefs}. The data was obtained from \cite{hernan2010causal} website, as it is used as a canonical example worked throughout the book. We followed it by controlling for demographics, smoking history and activity levels as confounders.

\section{Summary and discussion} 
Positivity is a necessary assumption for causal inference, ensuring treatment groups are comparable, so causal conclusions could be drawn.
Therefore, verifying it holds is an important part of the analysis process.
We present a scalable approach for detecting if positivity violations occur in a dataset and characterizing the subspaces these violations originate from. 

We do so by exploiting decision trees for the task.
We train them to differentiate the treatment groups,
so they divide the covariate space into disjoint regions, each region with maximized homogeneity in terms of treatment assignment.
If such homogeneous treatment regions are found, they are, by definition, positivity violating subspaces.

Since these regions are represented by the tree's leaves, we can characterize them intuitively, using covariate-level queries defined by the path along the decision tree.
Such simple portrayal is comparable with tabular methods for detecting positivity \citep{messer2010effects}.
However, in contrast to such methods, we can effortlessly scale to detect complex spaces, since detection is done algorithmically.
The tree representation can handle more variables than a table, and rules for each variable are chosen automatically so as to maximize discrimination, rather than being predefined by the user.


By correctly exploiting decision trees, our method does not trade-off between interpretability and scalability.

However, using off-the-shelf decision trees is not problem-free.
We discussed their tendency to overfit and linked it to their over-sensitivity and discovery of assumingly meaningless violations. 
We also suggested a range of ways to mitigate those false-discoveries using regularization, domain expert knowledge, probability and measuring how consistent violations are across an additional Random Forest. 

To summarize the findings in a concise way, we accompany the analysis above with a dedicated visualization of the decision tree. 
The visualization aim to highlight the most meaningful subspaces violating positivity.
It allows an easy overview of the scale and severity of violations in a dataset. 
The interactive version also allows a more thorough investigation into the subspaces to obtain additional information (see snapshot in Figure \ref{fig:suppfig1}).
Since the methodology is heuristic and threshold-based, the visualization can guide users toward choosing these hyperparameters.

Finally, we demonstrated the method on both synthetic and real-world examples. We showed how to interpret the visualization in both a positive example (lack of covariate overlap) and a negative (good covariate overlap) one. 

As data increase in complexity, so does the methods dealing with it should adjust to keep pace. 
Utilizing decision trees to act as interpretable detectors of biases in data can turn useful across a wide range of domains, in which treatment assignment in the context of causal inference is only one of them. 
We believe this method to be useful for causal inference researchers. We advocate incorporating it in workflows to help detect biases in the data prior to analyzing it, and therefore focusing on causal questions that can be supported by the data.

\section*{Acknowledgments}
The authors would like to thank the researchers of the Machine Learning for Healthcare group at IBM-Research Haifa Labs. Specifically, to Chen Yanover and Tal El-Hay for their thoughtful insights and help with background research.

\bibliographystyle{plain}
\bibliography{positivitree}

\clearpage
\newpage
\appendix
\section*{Supplementary figures}
\setcounter{figure}{0}  
\renewcommand{\thefigure}{S\arabic{figure}}

\begin{figure}[h]
  \includegraphics[width=\linewidth]{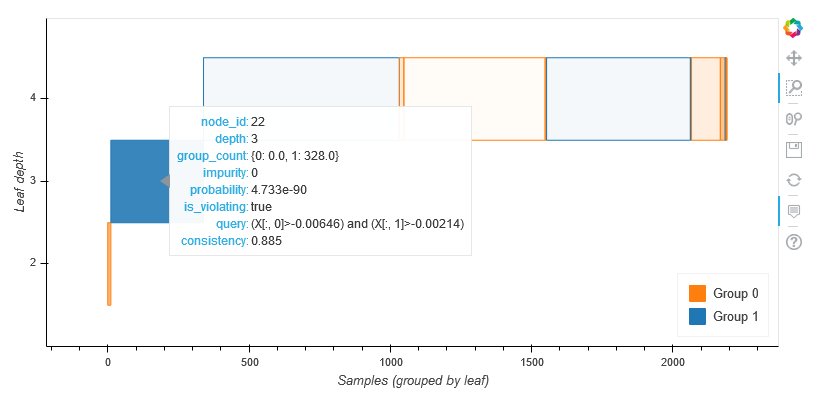}
  \caption{A snapshot of the exact same plot as Figure 1B, only its interactive version, where a box with additional information is shown upon hovering over the leaf (for details see Methods). The information is \emph{leaf depth} (the shallower the leaf, the simpler the corresponding subspace is, therefore violations there might be more important); the  \emph{number of samples from each group} (328 from group 1, and none from group 0);  \emph{impurity} (this case entropy, corresponding to the group counts);  \emph{probability} of the subspace, modeled with hypergeometric distribution shows a small value – meaning this violation is unlikely to happen by chance;  \emph{is\_violating} denotes whether the leaf counts as a violation by the tree;  \emph{aggregated consistency} value of the samples belonging to that leaf (a value being close to 1 means they were consistently flagged in the random forest, thus the opaque color of the rectangle); and finally, the data query characterizing the subspace corresponding to the leaf (in this case, defining the first quadrant, see Fig 1A).}
  \label{fig:suppfig1}
\end{figure}

\begin{figure}
  \includegraphics[width=\linewidth]{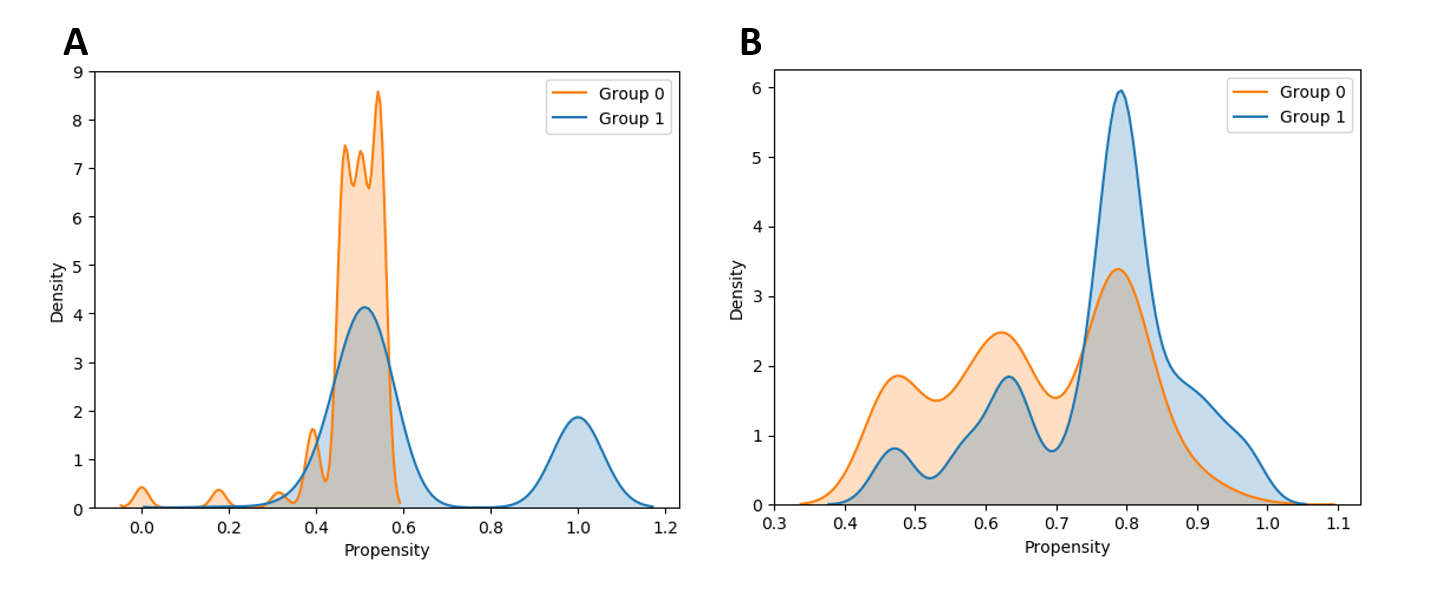}
  \caption{A more classic approach for detecting positivity violation for the two datasets presented in the main text, by examining the overlap of the propensity score distributions between the treatment groups. (\textbf{A}) is for the synthetic example showing a bulge in density around propensity = 1 exclusive for group 1, which suggest a violation does exist (Fig 1). Conversely, for the real-data example in (\textbf{B}) we see a good overlap of scores between groups, enforcing the fact we were not able to detect violations using the decision-tree and that the t-SNE projection showed good mixture of the groups (Fig 2). We note that this method may detect a problem exists, but characterizing its origin depends on the model used to obtain the scores.}
  \label{fig:suppfig2}
\end{figure}

\end{document}